\documentclass[11pt]{article}

\usepackage[preprint]{acl}

\usepackage{times}
\usepackage{latexsym}

\usepackage[T1,T5]{fontenc}

\usepackage[utf8]{inputenc}
\usepackage{microtype}
\usepackage{inconsolata}
\usepackage{graphicx}
\usepackage{booktabs}
\usepackage{multirow}
\usepackage{xcolor}
\usepackage{amsmath}

\title{How Well Do Decision Models Transfer to Vietnamese? An Empirical Study of Text Classification}

\author{
Phat Huu Ha$^{1,2}$ \and
Nguyen Xuan Phuc$^{1,2}$ \and
Nong Nhut Quy$^{1,2}$ \and \\
\textbf{Trinh Tran Tran$^{1,2}$} \textbf{\and }
\textbf{Nguyen Minh Triet$^{1,2}$} \textbf{\and} \\
\textbf{Dang Van Thin}$^{1,2}$
\\
$^1$University of Information Technology, Ho Chi Minh City, Vietnam \\
$^2$Vietnam National University HoChiMinh City, Vietnam\\
\texttt{Corresponding Author: thindv@uit.edu.vn}
}

\begin{document}
\maketitle

\begin{abstract}
Recent advances in specialized decision models have expanded the possibilities for zero-shot classification, yet their effectiveness on Vietnamese remains underexplored. This study evaluates four specialized decision models across 13 classification settings spanning six task groups, examining zero-shot performance, sensitivity to label-description language, and robustness to Vietnamese linguistic variations. The evaluation considers English, Vietnamese, and bilingual label descriptions, alongside common linguistic perturbations involving teencode, typographical errors, and diacritic removal. Under English label descriptions, JEV achieves the strongest overall performance, ranking first in 11 classification settings. The effects of label-description language vary across models and datasets, with bilingual descriptions providing no consistent improvement. Robustness experiments reveal that diacritic removal causes the most substantial performance degradation, reducing mean macro-$F_1$ by 9.51--15.47 points across model configurations, considerably more than teencode or typographical errors. These findings highlight the importance of label-description choices and Vietnamese orthographic variation when evaluating specialized decision models for zero-shot text classification.
\end{abstract}

\section{Introduction}
\label{sec:intro}

Text classification is a fundamental task in natural language processing, covering applications such as sentiment analysis, intent detection, and natural language inference. General-purpose large language models (LLMs) have demonstrated the ability to perform such tasks through zero-shot prompting. Recently, specialized decision models have emerged to provide structured predictions over predefined candidate options without free-form text generation. Models such as JEV supports classification through typed decision interfaces or user-defined schemas \citep{typesafe2026jev}.

Recent studies have evaluated specialized decision models across various tasks and domains. \citet{deusser2026jev} evaluate JEV across 37 datasets, reporting strong performance on several tasks alongside considerable variation in results. Other studies examine decision models and compare their performance with general-purpose LLMs \citep{duan2026generaldecision,ibrahim2026evaluatingdecisionmodels}. Despite these efforts, their performance on Vietnamese text classification remains insufficiently characterized, particularly across tasks with different label inventories and linguistic characteristics.

Classification performance may also depend on how labels are specified and how input text is written. Previous research has shown that option labels can influence predictions in typed decision models even when label definitions are provided \citep{azizi2026labels}. This motivates examining the effects of label-description language on Vietnamese classification tasks. Vietnamese user-generated text also contains abbreviations, nonstandard spellings, and omitted diacritics \citep{nguyen2024vilexnorm,nguyen-etal-2023-visobert}. Previous evaluations of Vietnamese pretrained language models have reported substantial performance degradation under diacritic removal \citep{nguyen-etal-2023-visobert}. These findings motivate further investigation of decision-model sensitivity to both label descriptions and Vietnamese linguistic variations.

This study evaluates four specialized decision models across 13 Vietnamese classification settings spanning six task groups, with additional comparisons against previously published LLM results. The evaluation examines zero-shot classification performance, the effects of label-description language, and robustness to Vietnamese linguistic perturbations. Under English label descriptions, JEV achieves the highest performance in 11 settings. The effects of label-description language vary across models and datasets, while diacritic removal causes the largest performance degradation among the evaluated perturbations.

The main contributions of this study are:
\begin{itemize}
    \item \textbf{Zero-shot evaluation of decision models for Vietnamese.} A systematic comparison across diverse classification tasks, identifying differences in model performance at both task and class levels.
    \item \textbf{Analysis of label-description sensitivity and linguistic robustness.} An investigation of the effects of label-description language and Vietnamese linguistic variations, revealing model-dependent behavior and substantial sensitivity to diacritic removal.
\end{itemize}

\section{Related Work}
\label{sec:related-work}

\subsection{Specialized Decision Models and Their Evaluation}

Specialized decision models produce structured predictions over predefined candidate options without free-form text generation. JEV and LAYA provide typed decision interfaces, while GLiNER2.5-multi-Decide supports multilingual classification with user-specified labels \citep{typesafe2026jev,convai2026laya,zaratiana2025gliner2efficientmultitaskinformation}. These systems support related classification tasks despite differences in their architectures and inference interfaces.

Recent evaluations show that decision-model performance varies across tasks and evaluation conditions. Broad benchmarks report competitive results on several classification and knowledge tasks, alongside limitations on more demanding tasks and settings \citep{deusser2026jev,duan2026generaldecision,ibrahim2026evaluatingdecisionmodels,li2026jevllm}. More focused studies reveal behavior not captured by aggregate scores. Experiments on legal inference identify sensitivity to request configuration, comparisons of JEV and LAYA show different performance and robustness profiles, and analyses of JEV report biases in the distribution of selected labels and ordinal scores \citep{zhang2026samescores,li2026fastmodels,gao2026ordinalbias}. However, evidence across diverse Vietnamese classification tasks remains limited.

\subsection{Label Specifications and Vietnamese Linguistic Robustness}

The specification of candidate labels can influence zero-shot classification. Studies of generative LLMs show that label wording and definitions affect classification performance \citep{mu2024promptcomplexity,li2025labelset}. Related findings have been reported for typed decision models, where option names can exert greater influence than their accompanying definitions \citep{azizi2026labels}. This evidence motivates examining the language of label descriptions as a separate factor in Vietnamese classification.

Robustness evaluation examines model behavior under controlled changes to input text. CheckList and TextFlint establish methods for testing linguistic transformations, while DecisionRobust extends perturbation-based evaluation to decision models and emphasizes the importance of validating that transformed inputs retain their original labels \citep{ribeiro2020checklist,wang2021textflint,lan2026decisionrobust}. For Vietnamese, ViLexNorm documents nonstandard lexical forms in social media, and ViSoBERT reports performance degradation under diacritic removal \citep{nguyen2024vilexnorm,nguyen-etal-2023-visobert}. These findings motivate evaluating the effects of Vietnamese linguistic variations on specialized decision models.

\section{Experimental Setup}
\label{sec:experimental-setup}

\subsection{Tasks and Datasets}
\label{sec:tasks-datasets}

The evaluation covers 13 classification settings from 12 Vietnamese datasets across six task groups: topic classification, emotion recognition, hate speech detection, sentiment analysis, intent detection, and natural language inference (NLI). The datasets include UIT-VSFC~\citep{vannguyen2018uitvsfc}, NEU-ESC~\citep{mai2025neuesc}, UIT-VSMEC~\citep{ho2019vsmec}, UIT-ViOCD~\citep{nhung2021viocd}, ViHSD~\citep{luu2021vihsd}, UIT-ViCTSD~\citep{nguyen2021victsd}, PhoATIS~\citep{dao2021phoatis}, ViMQ~\citep{huy2021vimq}, ViNLI~\citep{huynh2022vinli}, ViANLI~\citep{huynh2026vianli}, ViMedNLI~\citep{phan2023vimednli}, and vnNLI~\citep{anh2022nlivlsp}. UIT-VSFC is evaluated for both sentiment and topic classification.

\begin{table*}[t]
\centering
\footnotesize
\setlength{\tabcolsep}{1.2pt}
\renewcommand{\arraystretch}{1.12}
\caption{Zero-shot classification performance (\%) using English label descriptions and a supervised fine-tuned PhoBERT baseline. Acc, WF1, and MF1 denote accuracy, weighted-$F_1$, and macro-$F_1$. Bold and underlined MF1 indicate the highest and second-highest reported scores. Dashes denote unavailable values. Means cover the 12 settings shared by all four zero-shot models.}
\label{tab:main-results}
\begin{tabular*}{\textwidth}{@{\extracolsep{\fill}}l*{15}{c}@{}}
\toprule
\multicolumn{1}{@{}c}{\multirow{2}{*}{\textbf{Dataset}}}
& \multicolumn{3}{c}{\textbf{LAYA}}
& \multicolumn{3}{c}{\textbf{GLiNER}}
& \multicolumn{3}{c}{\textbf{Quyet}}
& \multicolumn{3}{c}{\textbf{JEV}}
& \multicolumn{3}{c@{}}{\textbf{Fine-tuning BERT}} \\
\cmidrule(lr){2-4}
\cmidrule(lr){5-7}
\cmidrule(lr){8-10}
\cmidrule(lr){11-13}
\cmidrule(l){14-16}
& \multicolumn{1}{c}{Acc}
& \multicolumn{1}{c}{WF1}
& \multicolumn{1}{c}{\textbf{MF1}}
& \multicolumn{1}{c}{Acc}
& \multicolumn{1}{c}{WF1}
& \multicolumn{1}{c}{\textbf{MF1}}
& \multicolumn{1}{c}{Acc}
& \multicolumn{1}{c}{WF1}
& \multicolumn{1}{c}{\textbf{MF1}}
& \multicolumn{1}{c}{Acc}
& \multicolumn{1}{c}{WF1}
& \multicolumn{1}{c}{\textbf{MF1}}
& \multicolumn{1}{c}{Acc}
& \multicolumn{1}{c}{WF1}
& \multicolumn{1}{c@{}}{\textbf{MF1}} \\
\midrule
\multicolumn{16}{@{}l}{\textit{Topic classification}} \\
NEU-ESC
& 26.49 & 30.02 & 14.79
& 62.30 & 58.80 & 27.10
& 47.19 & 49.19 & 23.08
& 51.79 & 55.83 & \underline{37.31}
& 78.65 & \multicolumn{1}{c}{--} & \textbf{62.73} \\
UIT-VSFC
& 49.15 & 57.08 & 46.02
& 44.91 & 55.03 & 48.99
& 82.56 & 81.86 & 63.63
& 83.77 & 84.61 & \underline{74.36}
& 88.22 & \multicolumn{1}{c}{--} & \textbf{78.08} \\
\midrule
\multicolumn{16}{@{}l}{\textit{Emotion recognition}} \\
UIT-VSMEC
& 34.34 & 35.80 & 31.89
& 18.76 & 6.13 & 5.10
& 56.28 & 56.59 & \underline{51.76}
& 55.70 & 54.58 & 49.99
& 64.65 & 63.80 & \textbf{59.06} \\
\midrule
\multicolumn{16}{@{}l}{\textit{Hate speech detection}} \\
ViHSD
& 58.95 & 65.38 & 38.35
& 10.33 & 1.98 & 6.25
& 29.12 & 37.45 & 30.59
& 83.70 & 83.87 & \underline{59.34}
& 86.75 & 86.52 & \textbf{64.76} \\
UIT-ViCTSD
& 87.80 & 85.61 & 58.08
& 87.30 & 83.88 & 50.90
& 89.50 & 88.84 & \underline{69.89}
& 89.50 & 88.73 & 69.32
& 90.78 & 90.27 & \textbf{71.31} \\
UIT-ViOCD
& 51.55 & 38.55 & 39.00
& 63.39 & 58.17 & 58.41
& 87.07 & 86.99 & 86.98
& 92.90 & 92.86 & \underline{92.85}
& 94.35 & 94.35 & \textbf{94.35} \\
\midrule
\multicolumn{16}{@{}l}{\textit{Sentiment analysis}} \\
UIT-VSFC
& 49.81 & 58.71 & 45.67
& 42.89 & 50.80 & 39.49
& 75.87 & 81.12 & 64.46
& 86.58 & 88.44 & \underline{73.44}
& 93.87 & 93.45 & \textbf{82.83} \\
\midrule
\multicolumn{16}{@{}l}{\textit{Intent detection}} \\
PhoATIS
& 15.79 & 17.94 & 19.86
& 23.74 & 26.29 & \underline{27.72}
& \multicolumn{1}{c}{--} & \multicolumn{1}{c}{--} & \multicolumn{1}{c}{--}
& 81.08 & 87.60 & \textbf{79.26}
& 90.82 & 90.82 & \multicolumn{1}{c}{--} \\
ViMQ
& 44.00 & 44.71 & 37.96
& 48.40 & 46.48 & 38.77
& 38.70 & 31.34 & 37.15
& 79.90 & 79.42 & \underline{80.20}
& \multicolumn{1}{c}{--} & \multicolumn{1}{c}{--} & \textbf{89.50} \\
\midrule
\multicolumn{16}{@{}l}{\textit{Natural language inference}} \\
ViNLI
& 47.61 & 46.84 & 46.66
& 28.28 & 23.86 & 23.63
& 51.19 & 48.69 & 48.48
& 79.91 & 80.85 & \textbf{81.00}
& 80.67 & \multicolumn{1}{c}{--} & \underline{80.69} \\
ViANLI
& 31.80 & 31.21 & 31.21
& 32.10 & 20.24 & 20.25
& 34.40 & 33.02 & \underline{33.01}
& 53.80 & 53.48 & \textbf{53.49}
& \multicolumn{1}{c}{--} & \multicolumn{1}{c}{--} & \multicolumn{1}{c}{--} \\
ViMedNLI
& 58.44 & 57.28 & 57.28
& 36.78 & 23.99 & 23.99
& 59.07 & 59.03 & 59.03
& 73.14 & 73.40 & \underline{73.40}
& 73.70 & \multicolumn{1}{c}{--} & \textbf{73.70} \\
vnNLI
& 71.99 & 71.79 & 71.79
& 40.24 & 32.20 & 32.19
& 75.34 & 75.24 & \underline{75.24}
& 85.71 & 85.89 & \textbf{85.90}
& 73.76 & \multicolumn{1}{c}{--} & 73.80 \\
\midrule
\textbf{Mean (12 shared)}
& 50.99 & 51.92 & 43.23
& 42.97 & 38.46 & 31.26
& 60.52 & 60.78 & \underline{53.61}
& 76.37 & 76.83 & \textbf{69.22}
& \multicolumn{1}{c}{--} & \multicolumn{1}{c}{--} & \multicolumn{1}{c}{--} \\
\bottomrule
\end{tabular*}
\end{table*}

\subsection{Models}
\label{sec:models}

Four specialized decision models are evaluated. JEV~\citep{typesafe2026jev} is accessed through its hosted API using the \texttt{typesafe/jev-1.13} model version. The remaining three models are evaluated locally using open-weight checkpoints: LAYA Multilingual (322M parameters)\footnote{\url{https://huggingface.co/convaiinnovations/laya-multilingual}}, GLiNER2.5-multi-Decide (287M parameters)\footnote{\url{https://huggingface.co/fastino/GLiNER2.5-multi-Decide}}, and Quyet-1.0-Small (328M parameters)\footnote{\url{https://huggingface.co/chinhnc/Quyet-1.0-Small}}. Quyet supports a maximum of ten candidate labels and is therefore excluded from PhoATIS, which contains 28 intent labels.

\subsection{Zero-Shot Evaluation Protocol}
\label{sec:inference-protocol}

All models are evaluated with frozen weights, without task-specific fine-tuning or labeled demonstrations. Each model receives the task input and its predefined candidate labels. To examine the effects of label-description language, three configurations are considered: English (EN), Vietnamese (VI), and bilingual English--Vietnamese (EN+VI). Label identities and their order remain fixed across configurations, while EN+VI combines the English and Vietnamese descriptions. English descriptions serve as the reference configuration for the main evaluation.

Macro-$F_1$ is the primary evaluation metric, with accuracy and weighted-$F_1$ reported as supplementary measures. Macro-$F_1$ is computed over classes represented in the gold test set. Robustness is measured by the change in macro-$F_1$ between original and perturbed inputs under the same model configuration. The perturbation procedures are described in Section~\ref{sec:linguistic-robustness}.

\begin{figure*}[t]
\centering
\includegraphics[width=\textwidth]{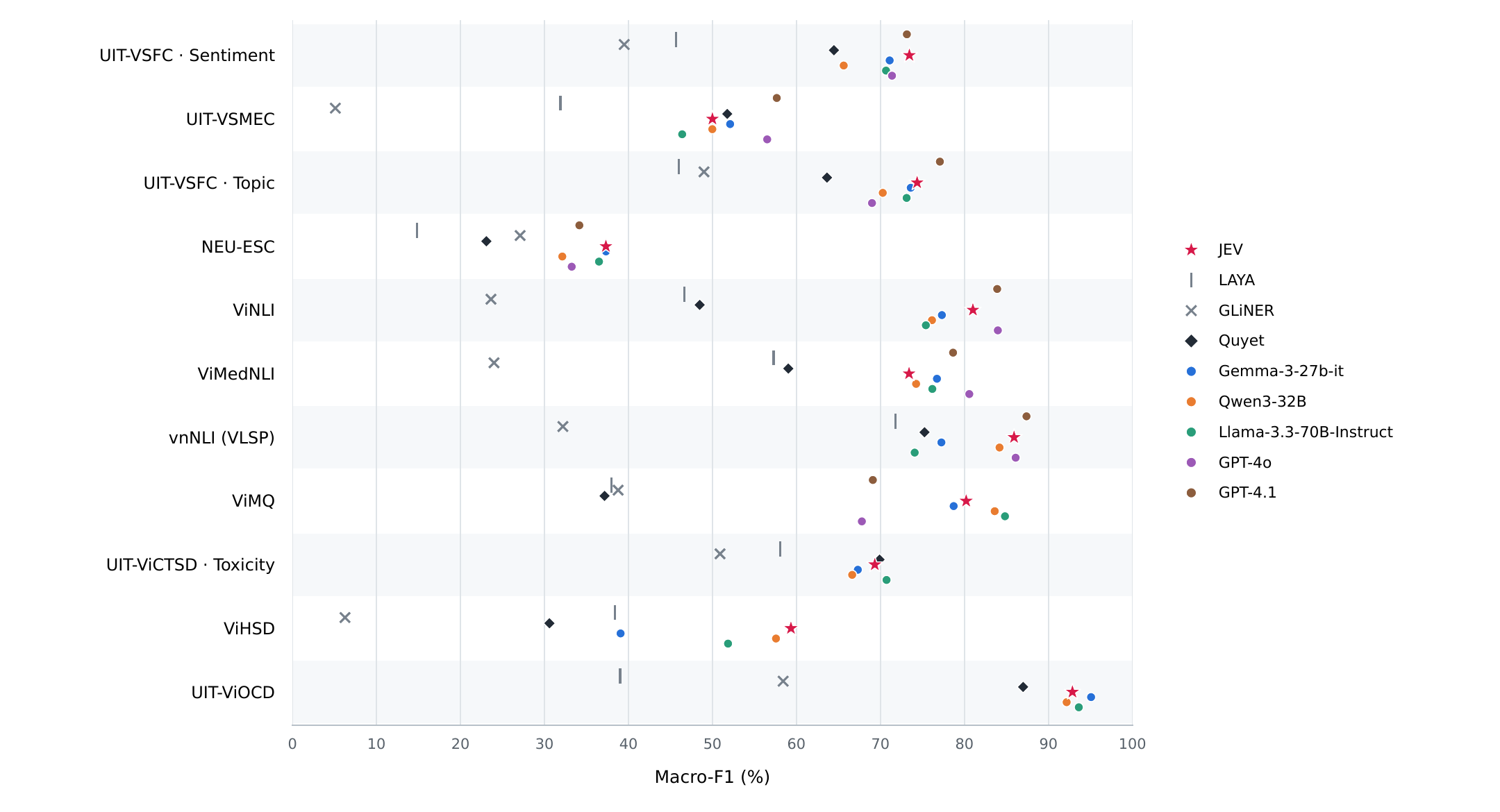}
\caption{Zero-shot macro-$F_1$ (\%) of specialized decision models and general-purpose LLMs across 11 Vietnamese classification settings.}
\label{fig:decision-llm-comparison}
\end{figure*}

\section{Results and Analysis}
\label{sec:results}

\subsection{Overall Performance Across Tasks}
\label{sec:overall-performance}

Table~\ref{tab:main-results} presents the zero-shot classification results under English label descriptions. JEV achieves the highest macro-$F_1$ in 11 of the 13 settings, with an average of 69.22 across the 12 settings shared by all models, followed by Quyet (53.61), LAYA (43.23), and GLiNER (31.26). Its advantage is particularly pronounced in intent classification, achieving 79.26 macro-$F_1$ on PhoATIS and 80.20 on ViMQ, exceeding the next-best models by 51.54 and 41.43 points, respectively. Notably, PhoATIS contains 28 candidate intent classes, indicating strong performance even with a relatively large label inventory. Substantial variation is also observed within task families. In topic classification, JEV achieves 74.36 macro-$F_1$ on UIT-VSFC but only 37.31 on NEU-ESC. Across the NLI benchmarks, all four models obtain their lowest scores on ViANLI, a dataset constructed using adversarial examples to challenge existing NLI systems. Nevertheless, JEV does not consistently outperform the other models, with Quyet achieving slightly higher scores on UIT-VSMEC and UIT-ViCTSD. Differences are also observed across evaluation metrics. On NEU-ESC, GLiNER achieves the highest accuracy (62.30\%), whereas JEV obtains the highest macro-$F_1$ (37.31\%), illustrating how model rankings can differ depending on the evaluation metric.

\subsection{Comparison with LLMs}
\label{sec:llm-comparison}

To contextualize the performance of specialized decision models, we compare their zero-shot classification results with those reported for five general-purpose LLMs in prior work \cite{thin-etal-2024-prompt}. Figure~\ref{fig:decision-llm-comparison} presents macro-$F_1$ scores on 11 overlapping classification settings, with detailed results provided in Table~\ref{tab:decision-llm-comparison} in the appendix. The comparison uses English label descriptions for the decision models and previously reported zero-shot prompting results for the LLMs.

JEV demonstrates competitive performance against substantially larger general-purpose models, outperforming all three open-weight LLMs on five of the 11 settings: UIT-VSFC topic and sentiment classification, ViHSD, ViNLI, and vnNLI. It also achieves the highest reported macro-$F_1$ on ViHSD (59.34), exceeding Qwen3-32B (57.56), and on UIT-VSFC sentiment classification (73.44), slightly surpassing GPT-4.1 (73.14). On NEU-ESC, JEV performs comparably to the strongest LLM, Gemma-3-27b-it (37.31 versus 37.32). Among the remaining decision models, Quyet is particularly competitive on UIT-ViCTSD, reaching 69.89 macro-$F_1$ compared with 70.72 for Llama-3.3-70B-Instruct. Nevertheless, the relative performance varies across tasks. On UIT-VSMEC, GPT-4.1 exceeds JEV by 7.66 macro-$F_1$ points, while GPT-4o achieves a 7.17-point advantage on ViMedNLI. Llama-3.3-70B-Instruct also outperforms JEV on ViMQ by 4.62 points.

\subsection{Error Analysis}
\label{sec:error-analysis}

On PhoATIS, JEV frequently predicts compound intents for queries annotated with a single intent. Of its 169 incorrect predictions, 140 include the gold intent together with an additional intent. The most frequent confusion is \texttt{flight} $\rightarrow$ \texttt{flight\#flight\_time} (109 cases). For example, JEV assigns \texttt{flight\#flight\_time} to the query ``cho tôi biết các chuyến bay từ thành phố hà nội đến thanh hóa khởi hành sau 5 giờ chiều thứ tư'' (``show me flights from Hanoi to Thanh Hoa departing after 5 p.m. on Wednesday''), although its gold label is \texttt{flight}. The departure time specifies a search constraint, not a separate request for flight-time information. JEV correctly identifies 19 of the 20 examples annotated with compound intents, suggesting that distinguishing requested information from query constraints is a prominent difficulty in this setting. However, the small number of compound examples limits conclusions about compound-intent recognition more broadly.

\begin{figure}[t]
    \centering
    \includegraphics[width=\columnwidth]{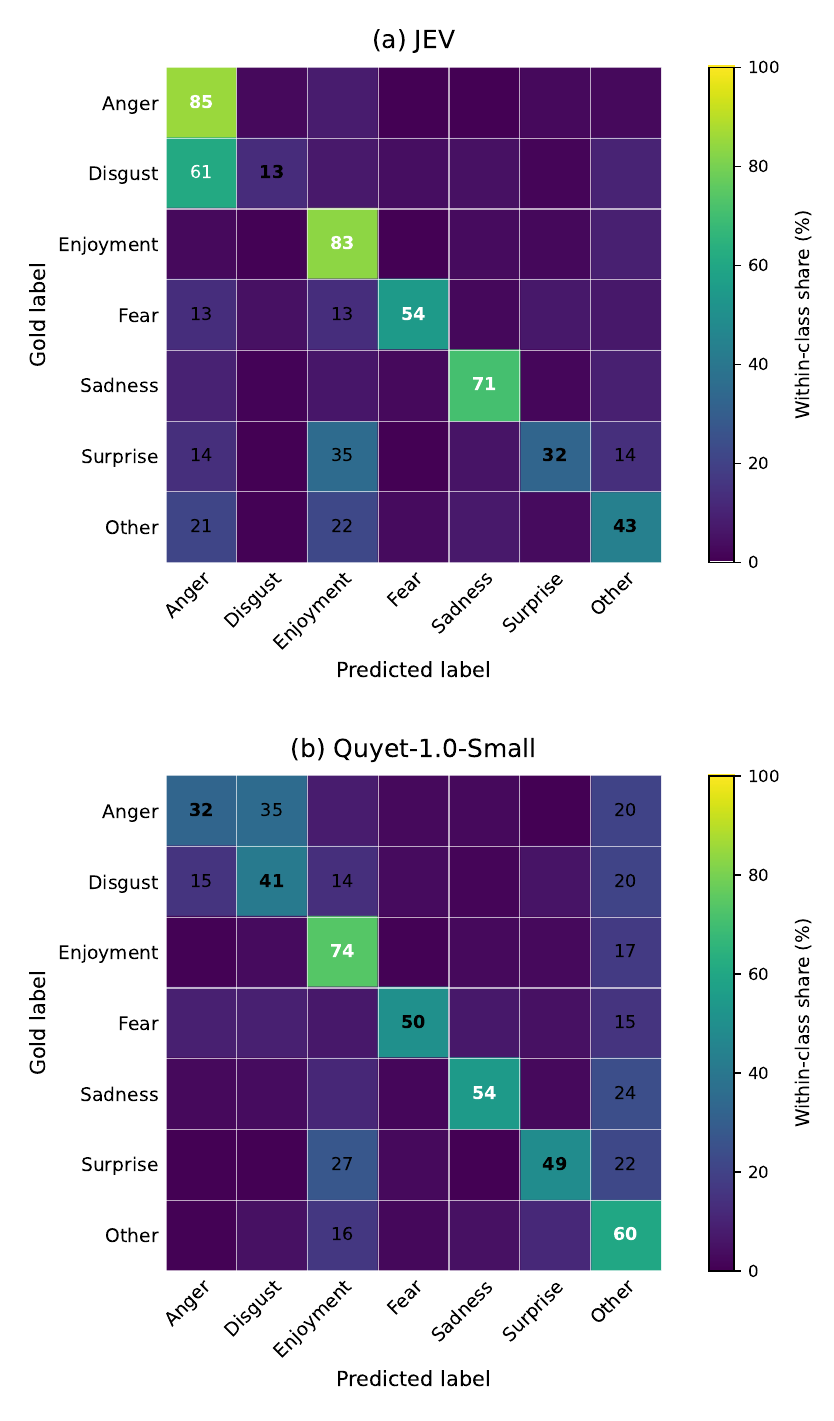}
    \caption{Comparison of JEV and Quyet confusion matrices on UIT-VSMEC.}
    \label{fig:vsmec-confusions}
\end{figure}

On UIT-VSMEC, a prominent difference between JEV and Quyet concerns the distinction between \textit{anger} and \textit{disgust}. JEV misclassifies 80 of the 132 \textit{disgust} examples as \textit{anger}, obtaining an $F_1$ of 21.94 for \textit{disgust}, compared with 49.09 for Quyet. Conversely, JEV achieves higher recall for \textit{anger} (85.00\% versus 32.50\%). Figure~\ref{fig:vsmec-confusions} shows these contrasting confusion profiles, which are not reflected in the relatively close overall macro-$F_1$ scores.

Another recurring error pattern is the disproportionate selection of particular labels. On UIT-VSFC sentiment classification, all four models overpredict \textit{neutral}, assigning it to 395--1,875 examples against 167 gold instances, with neutral precision ranging from 7.36\% to 27.09\%. For example, the positive comment ``trình bày vấn đề một cách chính xác và dễ hiểu'' (``presents the material accurately and clearly'') is classified as \textit{neutral} by JEV, LAYA, and GLiNER. On ViNLI, 551 of JEV's 601 errors involve predicting \textit{neutral} for examples with other gold labels. More extreme prediction concentration occurs for GLiNER, which assigns \textit{other} to 692 of 693 UIT-VSMEC examples and \textit{hate} to 6,678 of 6,680 ViHSD examples. These distributions indicate that, in some evaluated configurations, errors are dominated by repeated selection of particular labels rather than confusions spread across multiple categories.

\begin{table*}[t]
    \centering
    \caption{Changes in macro-$F_1$ relative to English label descriptions (percentage points).}
    \label{tab:label-language}
    \footnotesize
    \setlength{\tabcolsep}{4pt}
    \renewcommand{\arraystretch}{1.08}
    \begin{tabular}{l*{8}{c}}

        \toprule
        \multicolumn{1}{c}{\multirow{2}{*}{\textbf{Dataset}}}
        & \multicolumn{2}{c}{\textbf{LAYA}}
        & \multicolumn{2}{c}{\textbf{GLiNER}}
        & \multicolumn{2}{c}{\textbf{Quyet}}
        & \multicolumn{2}{c}{\textbf{JEV}} \\
        \cmidrule(lr){2-3}
        \cmidrule(lr){4-5}
        \cmidrule(lr){6-7}
        \cmidrule(lr){8-9}
        & \multicolumn{1}{c}{VI}
        & \multicolumn{1}{c}{EN+VI}
        & \multicolumn{1}{c}{VI}
        & \multicolumn{1}{c}{EN+VI}
        & \multicolumn{1}{c}{VI}
        & \multicolumn{1}{c}{EN+VI}
        & \multicolumn{1}{c}{VI}
        & \multicolumn{1}{c}{EN+VI} \\
        \midrule

        \multicolumn{9}{l}{\textit{Topic classification}} \\
        NEU-ESC
        & $+6.06$ & $-4.07$
        & $-1.37$ & $+1.32$
        & $+3.63$ & $-3.18$
        & $-0.67$ & $-0.32$ \\
        UIT-VSFC
        & $-5.24$ & $-15.08$
        & $-11.65$ & $-13.68$
        & $+2.96$ & $-1.42$
        & $+0.26$ & $+0.84$ \\
        \midrule

        \multicolumn{9}{l}{\textit{Emotion recognition}} \\
        UIT-VSMEC
        & $-2.36$ & $-4.83$
        & $+0.57$ & $+0.59$
        & $+3.38$ & $-11.86$
        & $+1.42$ & $+1.55$ \\
        \midrule

        \multicolumn{9}{l}{\textit{Hate speech detection}} \\
        ViHSD
        & $+0.68$ & $-4.02$
        & $-0.01$ & $-0.01$
        & $-14.23$ & $-18.94$
        & $-0.62$ & $+0.34$ \\
        UIT-ViCTSD
        & $-1.70$ & $-0.69$
        & $-3.31$ & $-2.23$
        & $-1.15$ & $-3.86$
        & $+0.14$ & $+0.77$ \\
        UIT-ViOCD
        & $+13.33$ & $-4.25$
        & $-9.64$ & $-23.48$
        & $-1.93$ & $-19.50$
        & $-0.19$ & $+0.37$ \\
        \midrule

        \multicolumn{9}{l}{\textit{Sentiment analysis}} \\
        UIT-VSFC
        & $+8.00$ & $+5.28$
        & $-4.43$ & $-23.44$
        & $-6.66$ & $-11.18$
        & $-1.32$ & $-0.22$ \\
        \midrule

        \multicolumn{9}{l}{\textit{Intent detection}} \\
        PhoATIS
        & $+2.14$ & $0.00$
        & $-4.68$ & $-3.41$
        & -- & --
        & $-4.82$ & $-1.09$ \\
        ViMQ
        & $-2.37$ & $-18.70$
        & $-22.46$ & $-22.06$
        & $-0.86$ & $-7.24$
        & $-0.48$ & $+1.15$ \\
        \midrule

        \multicolumn{9}{l}{\textit{Natural language inference}} \\
        ViNLI
        & $+4.56$ & $-1.16$
        & $-13.85$ & $-13.85$
        & $+21.06$ & $+15.97$
        & $+1.50$ & $+0.07$ \\
        ViANLI
        & $-0.55$ & $-0.23$
        & $-2.61$ & $-2.68$
        & $-1.65$ & $+0.05$
        & $-0.12$ & $-0.77$ \\
        ViMedNLI
        & $-2.67$ & $-4.78$
        & $-6.45$ & $-6.45$
        & $+1.03$ & $-3.87$
        & $-0.67$ & $-0.81$ \\
        vnNLI
        & $-0.92$ & $-1.27$
        & $-12.74$ & $-12.01$
        & $-0.29$ & $-0.88$
        & $-0.73$ & $-0.33$ \\
        \midrule

        \textbf{Mean (12 shared)}
        & $+1.40$ & $-4.48$
        & $-7.33$ & $-9.83$
        & $+0.44$ & $-5.49$
        & $-0.12$ & $+0.22$ \\
        \bottomrule
    \end{tabular}
\end{table*}

\subsection{Effects of Label-Description Language}
\label{sec:label-description-language}

Table~\ref{tab:label-language} shows distinct patterns of sensitivity to label-description language across models. JEV exhibits the most stable average performance, with macro-$F_1$ varying by only 0.35 points across the three configurations and reaching its highest value under EN+VI descriptions (69.44). Quyet and LAYA achieve their highest average scores with Vietnamese descriptions, although the improvements over English are modest and both models perform substantially worse with bilingual descriptions. GLiNER, in contrast, performs best with English descriptions and exhibits the largest average degradation under VI and EN+VI. These results suggest that multilingual label descriptions do not necessarily benefit models built on multilingual encoders.

The effects are also strongly task-dependent. Quyet, for example, gains 21.06 macro-$F_1$ points on ViNLI with Vietnamese descriptions, largely through improved discrimination between \textit{other} and \textit{neutral}. However, the same configuration reduces its performance on ViHSD by 14.23 points, accompanied by increased misclassification of \textit{clean} examples as \textit{offensive}. 
Additional instance-level prediction comparisons are provided in Appendix~\ref{app:label-sensitivity}. This contrast suggests that the effect of description language is closely related to the distinctions between task-specific labels, rather than a general preference for Vietnamese descriptions. The degradation observed for LAYA and GLiNER is particularly pronounced in some settings. LAYA loses 18.70 macro-$F_1$ points on ViMQ with bilingual descriptions, while GLiNER assigns \textit{other} to every ViNLI example under both Vietnamese and bilingual descriptions. These findings indicate that label-description language affects not only classification performance but also the distribution of predicted labels, with substantial differences across models and tasks.

\section{Robustness to Vietnamese Linguistic Variation}
\label{sec:linguistic-robustness}

\subsection{Perturbation Design}
\label{sec:perturbation-design}

We consider three types of Vietnamese linguistic variation. \textit{Teencode} introduces informal lexical substitutions commonly found in online communication, such as ``không'' $\rightarrow$ ``ko'' (``not''). \textit{Typographical noise} simulates typing errors through Telex-related substitutions, diacritic modifications, and character-level edits. \textit{Diacritic removal} removes Vietnamese tone and vowel diacritics, including the conversion of \textit{đ} to \textit{d}. Each perturbation is applied independently to the original test inputs while retaining the original labels. For NLI, teencode and diacritic removal affect both the premise and hypothesis, whereas typographical noise is applied only to the hypothesis.

Table~\ref{tab:perturbation-audit} summarizes perturbation coverage and character edit burden across the 13 evaluation settings, with dataset-level statistics provided in Appendix~\ref{app:perturbation-config}. Coverage measures the proportion of modified inputs, while edit burden is the total character-level Levenshtein distance normalized by the original input length. Diacritic removal produces the highest coverage and edit burden, followed by typographical noise and teencode.

\begin{table}[t]
\centering
\small
\setlength{\tabcolsep}{5pt}
\caption{Perturbation coverage and character edit burden across the 13 evaluation settings.}
\label{tab:perturbation-audit}
\begin{tabular}{@{}lrr@{}}
\toprule
Perturbation & Changed (\%) & Edit burden (\%) \\
\midrule
Teencode & 61.70 & 2.14 \\
Typo & 90.18 & 4.93 \\
No accents & 98.10 & 18.28 \\
\bottomrule
\end{tabular}
\end{table}

\subsection{Performance Under Linguistic Perturbations}
\label{sec:perturbation-results}

\begin{table*}[t]
\centering
\footnotesize
\setlength{\tabcolsep}{2.1pt}
\renewcommand{\arraystretch}{1.10}
\caption{Changes in macro-$F_1$ under linguistic perturbations relative to original inputs (percentage points).}
\label{tab:perturbation-results}
\begin{tabular}{@{}l*{12}{r}@{}}
\toprule
\multicolumn{1}{@{}c}{\multirow{2}{*}{\textbf{Dataset}}}
& \multicolumn{3}{c}{\textbf{LAYA}}
& \multicolumn{3}{c}{\textbf{GLiNER}}
& \multicolumn{3}{c}{\textbf{Quyet}}
& \multicolumn{3}{c@{}}{\textbf{JEV}} \\
\cmidrule(lr){2-4}
\cmidrule(lr){5-7}
\cmidrule(lr){8-10}
\cmidrule(l){11-13}
& \multicolumn{1}{c}{NoAcc}
& \multicolumn{1}{c}{Teen}
& \multicolumn{1}{c}{Typo}
& \multicolumn{1}{c}{NoAcc}
& \multicolumn{1}{c}{Teen}
& \multicolumn{1}{c}{Typo}
& \multicolumn{1}{c}{NoAcc}
& \multicolumn{1}{c}{Teen}
& \multicolumn{1}{c}{Typo}
& \multicolumn{1}{c}{NoAcc}
& \multicolumn{1}{c}{Teen}
& \multicolumn{1}{c@{}}{Typo} \\
\midrule

\multicolumn{13}{@{}l}{\textit{Topic classification}} \\
NEU-ESC
& $-9.93$ & $-0.31$ & $-3.42$
& $-12.74$ & $-0.10$ & $-2.30$
& $-9.68$ & $+0.02$ & $-1.38$
& $-2.48$ & $-0.38$ & $-0.62$ \\
UIT-VSFC
& $-27.08$ & $-0.59$ & $-4.31$
& $-46.60$ & $+0.50$ & $-7.86$
& $-29.42$ & $+0.08$ & $-4.08$
& $-13.22$ & $+0.36$ & $-3.36$ \\
\midrule

\multicolumn{13}{@{}l}{\textit{Emotion recognition}} \\
UIT-VSMEC
& $-15.85$ & $-1.74$ & $-6.85$
& $-0.62$ & $-0.62$ & $+0.59$
& $-26.38$ & $-4.93$ & $-9.80$
& $-18.64$ & $-5.13$ & $-6.43$ \\
\midrule

\multicolumn{13}{@{}l}{\textit{Hate speech detection}} \\
ViHSD
& $-8.50$ & $-0.45$ & $-3.41$
& $-0.02$ & $-0.01$ & $0.00$
& $-5.93$ & $-0.34$ & $-1.38$
& $-16.07$ & $-2.39$ & $-4.97$ \\
UIT-ViCTSD
& $-2.75$ & $-0.39$ & $-1.40$
& $-3.81$ & $+1.41$ & $-1.95$
& $-15.83$ & $+0.29$ & $-4.71$
& $-8.23$ & $-0.62$ & $-2.60$ \\
UIT-ViOCD
& $+1.88$ & $+2.67$ & $+9.58$
& $-25.44$ & $-6.28$ & $-8.97$
& $-24.79$ & $-1.89$ & $-6.56$
& $-11.49$ & $-0.71$ & $-4.84$ \\
\midrule

\multicolumn{13}{@{}l}{\textit{Sentiment analysis}} \\
UIT-VSFC
& $-13.36$ & $-1.98$ & $-3.23$
& $-34.12$ & $-4.12$ & $-8.85$
& $-14.61$ & $-0.62$ & $-3.26$
& $-12.39$ & $-1.02$ & $-4.11$ \\
\midrule

\multicolumn{13}{@{}l}{\textit{Intent detection}} \\
PhoATIS
& $-19.27$ & $+0.63$ & $-3.53$
& $-22.51$ & $-0.46$ & $-5.74$
& -- & -- & --
& $-12.04$ & $+1.46$ & $-3.67$ \\
ViMQ
& $-12.54$ & $+1.63$ & $-1.92$
& $-21.98$ & $+2.66$ & $-4.34$
& $-5.08$ & $+1.78$ & $-3.61$
& $-4.94$ & $-0.39$ & $-2.07$ \\
\midrule

\multicolumn{13}{@{}l}{\textit{Natural language inference}} \\
ViNLI
& $-21.76$ & $+1.68$ & $-4.10$
& $-11.62$ & $-1.51$ & $-4.65$
& $-12.55$ & $+1.71$ & $-3.19$
& $-6.07$ & $-0.57$ & $-1.61$ \\
ViANLI
& $-4.67$ & $+1.68$ & $-1.17$
& $-2.51$ & $-0.99$ & $-2.01$
& $-5.73$ & $-0.24$ & $-2.00$
& $-7.95$ & $+0.27$ & $-0.66$ \\
ViMedNLI
& $-19.05$ & $-5.34$ & $-6.20$
& $-4.67$ & $-1.09$ & $-1.53$
& $-11.19$ & $-0.79$ & $-6.22$
& $-9.15$ & $-1.78$ & $-4.01$ \\
vnNLI
& $-31.62$ & $-3.13$ & $-8.01$
& $-12.91$ & $-1.92$ & $-5.43$
& $-24.48$ & $-0.76$ & $-8.31$
& $-3.46$ & $-0.85$ & $-1.50$ \\
\midrule

\textbf{Mean (12 shared)}
& $-13.77$ & $-0.52$ & $-2.87$
& $-14.75$ & $-1.01$ & $-3.94$
& $-15.47$ & $-0.47$ & $-4.54$
& $-9.51$ & $-1.10$ & $-3.07$ \\
\bottomrule
\end{tabular}
\end{table*}

Table~\ref{tab:perturbation-results} reports changes in macro-$F_1$ relative to unperturbed inputs. For each model, we use the label-description configuration that achieves the highest mean macro-$F_1$ on the original test data (EN+VI for JEV, EN for GLiNER, and VI for LAYA and Quyet) and keep it fixed across all perturbation conditions. JEV maintains the strongest overall performance across all three perturbation types. Its mean macro-$F_1$ across the 12 shared settings decreases from 69.44 on the original inputs to 59.93 under diacritic removal, 68.34 under teencode, and 66.37 under typographical noise, remaining above the other models in every condition. JEV achieves the highest macro-$F_1$ in all 13 settings under diacritic removal and in 12 of 13 settings under each of the other two perturbations. Its lead over the next-best model, Quyet, also widens from 15.39 points on the original inputs to 21.35 points under diacritic removal.

Diacritic removal produces the largest average performance loss for every model, with decreases in 47 of the 48 model--setting comparisons shared by all four models. JEV exhibits the smallest mean reduction (9.51 points), followed by LAYA (13.77), GLiNER (14.75), and Quyet (15.47). The effects nevertheless vary substantially across tasks. On UIT-VSFC topic classification, GLiNER loses 46.60 points and predicts \textit{others} for 3,164 of 3,166 instances; on the sentiment task, it predicts \textit{neutral} for 3,112 instances, accompanied by a 34.12-point decline. Differences are also pronounced on vnNLI, where LAYA and Quyet lose 31.62 and 24.48 points, respectively, compared with only 3.46 for JEV. Prediction-level comparisons further show that diacritic removal changes 19.57\% of JEV's decisions on average, compared with 40.60\% for Quyet and 54.29\% for LAYA. GLiNER exhibits a similarly low flip rate (19.83\%), but its predictions remain concentrated on particular labels in several settings. Detailed prediction transitions are reported in Appendix~\ref{app:noise-predictions}.

Teencode and typographical noise have smaller average effects, with mean reductions of 0.47--1.10 and 2.87--4.54 points, respectively. Although JEV retains the highest absolute performance, it does not consistently exhibit the smallest degradation: Quyet has the lowest mean loss under teencode, while LAYA is least affected by typographical noise. Notably, LAYA improves on UIT-ViOCD under all three perturbations, gaining 9.58 points under typographical noise. This improvement coincides with a reduction in its preference for \textit{non\_complaint}, from 480 of 549 predictions on the original inputs to 291 under typographical noise, closer to the 270 gold instances. Conversely, teencode causes notable declines on UIT-VSMEC for JEV (5.13 points) and Quyet (4.93 points). A targeted analysis of the Vietnamese negation marker \textit{không} in Appendix~\ref{app:negation-noise} further identifies differences in models' sensitivity to its informal written forms.

\section{Conclusion}
\label{sec:conclusion}

We presented a systematic evaluation of four specialized decision models for Vietnamese zero-shot text classification across 13 settings spanning six task families. Beyond overall classification performance, we examined the effects of label-description language, class-level prediction patterns, and robustness to Vietnamese linguistic variation. JEV achieved the highest macro-$F_1$ in 11 of the 13 settings under English label descriptions and maintained the strongest average performance across all three linguistic perturbations. However, model behavior varied considerably across tasks and evaluation conditions. Label-description language produced model- and task-dependent effects, with bilingual descriptions offering no consistent advantage. Diacritic removal caused the largest average performance degradation for all four models, although JEV exhibited the smallest mean reduction under this condition. These findings demonstrate that strong zero-shot performance does not necessarily imply consistent behavior across label configurations and linguistic variations, highlighting the importance of evaluating specialized decision models beyond aggregate benchmark scores.

\bibliography{custom}

\appendix

\section{Prediction-Level Sensitivity}
\label{app:prediction-sensitivity}

We examine instance-level prediction changes under two evaluation conditions: variations in label-description language and linguistic perturbations of the input text. A prediction flip occurs when the predicted label differs between paired configurations. W$\rightarrow$C, C$\rightarrow$W, and W$\rightarrow$W denote transitions from incorrect to correct, correct to incorrect, and between two incorrect predictions, respectively.

\subsection{Label-Description Language}
\label{app:label-sensitivity}

Table~\ref{tab:description-flips} presents selected prediction changes underlying the macro-$F_1$ differences reported in Section~\ref{sec:label-description-language}. On ViNLI, switching from English to Vietnamese descriptions changes 123 predictions for JEV but 930 for Quyet. For Quyet, this change corrects 682 previously incorrect predictions while introducing 151 new errors, consistent with its substantial improvement in macro-$F_1$. The effect is reversed on ViHSD, where the same change introduces 1,247 errors and corrects only 147. For LAYA, switching from Vietnamese to bilingual descriptions on ViMQ produces 393 newly incorrect predictions compared with 88 corrections. GLiNER also exhibits substantial prediction changes on ViNLI, with 987 of its 2,266 flips occurring between two incorrect labels.

\begin{table*}[t]
\centering
\footnotesize
\setlength{\tabcolsep}{4pt}
\renewcommand{\arraystretch}{1.10}
\caption{Selected instance-level prediction changes across label-description configurations.}
\label{tab:description-flips}
\begin{tabular}{@{}lllrrrrr@{}}
\toprule
Model & Dataset & Description change
& Flips / $n$ (\%)
& W$\rightarrow$C
& C$\rightarrow$W
& W$\rightarrow$W
& $\Delta$MF1 \\
\midrule
LAYA & ViMQ & VI $\rightarrow$ EN+VI
& 760 / 1,000 (76.00)
& 88 & 393 & 279 & $-16.33$ \\
\midrule
GLiNER & ViNLI & EN $\rightarrow$ VI
& 2,266 / 2,991 (75.76)
& 580 & 699 & 987 & $-13.85$ \\
\midrule
Quyet & ViNLI & EN $\rightarrow$ VI
& 930 / 2,991 (31.09)
& 682 & 151 & 97 & $+21.06$ \\
Quyet & ViHSD & EN $\rightarrow$ VI
& 1,555 / 6,680 (23.28)
& 147 & 1,247 & 161 & $-14.23$ \\
\midrule
JEV & ViNLI & EN $\rightarrow$ VI
& 123 / 2,991 (4.11)
& 84 & 28 & 11 & $+1.50$ \\
\bottomrule
\end{tabular}
\end{table*}

\subsection{Linguistic Perturbations}
\label{app:noise-predictions}

Table~\ref{tab:noise-transitions} summarizes prediction changes under diacritic removal, the perturbation producing the largest performance degradation in Section~\ref{sec:perturbation-results}. Results are averaged across the 12 shared evaluation settings, with each setting weighted equally. JEV exhibits considerably fewer prediction changes under diacritic removal than Quyet and LAYA. In particular, 11.73\% of its predictions change from correct to incorrect, compared with 23.33\% for Quyet and 27.95\% for LAYA. Despite a similarly low flip rate (19.83\%), GLiNER achieves substantially lower classification performance, with its predictions dominated by a single label in several settings. Its apparent prediction stability therefore partly reflects persistent label concentration.

\begin{table}[t]
\centering
\footnotesize
\setlength{\tabcolsep}{5pt}
\renewcommand{\arraystretch}{1.10}
\caption{Instance-level prediction transitions under diacritic removal (\%).}
\label{tab:noise-transitions}
\begin{tabular}{@{}lrrrr@{}}
\toprule
Model & Flips & W$\rightarrow$C
& C$\rightarrow$W & W$\rightarrow$W \\
\midrule
LAYA   & 54.29 & 12.15 & 27.95 & 14.19 \\
GLiNER & 19.83 & 3.33 & 12.29 & 4.21 \\
Quyet  & 40.60 & 8.13 & 23.33 & 9.15 \\
JEV    & 19.57 & 4.21 & 11.73 & 3.63 \\
\bottomrule
\end{tabular}
\end{table}

\subsection{Negation-Cue Sensitivity under Teencode}
\label{app:negation-noise}

We further examine the sensitivity of NLI predictions to informal variants of the Vietnamese negation marker \textit{không}. We identify 194 instances across the four NLI benchmarks in which \textit{không} occurs in the hypothesis but not the premise and is replaced by a teencode variant, such as ``ko'' or ``k'', without modifying the remaining input text. Fixed expressions in which \textit{không} does not function as negation, such as ``không khí'', ``không gian'', and ``hàng không'', are excluded. Replacing \textit{không} substantially reduces the proportion of \textit{contradiction} predictions for LAYA and GLiNER, by 20.62 and 14.43 percentage points, respectively. The corresponding changes are considerably smaller for JEV (3.09 points) and Quyet (2.06 points). LAYA and GLiNER also experience larger accuracy reductions on these instances. These results indicate greater sensitivity to the surface realization of negation cues in LAYA and GLiNER, even when the remaining input text is unchanged.

\begin{table}[t]
\centering
\footnotesize
\setlength{\tabcolsep}{5pt}
\renewcommand{\arraystretch}{1.10}
\caption{Effects of rewriting \textit{không} on NLI predictions (194 paired instances).}
\label{tab:negation-teencode}
\begin{tabular}{@{}lrrrr@{}}
\toprule
& \multicolumn{2}{c}{\textbf{Contradiction (\%)}} 
& \multicolumn{2}{c}{\textbf{Accuracy (\%)}} \\
\cmidrule(lr){2-3}
\cmidrule(l){4-5}
Model & Original & Teencode & Original & Teencode \\
\midrule
LAYA   & 52.06 & 31.44 & 67.01 & 44.85 \\
GLiNER & 23.71 & 9.28  & 35.05 & 24.23 \\
Quyet  & 58.76 & 56.70 & 74.23 & 70.10 \\
JEV    & 50.00 & 46.91 & 82.99 & 78.35 \\
\bottomrule
\end{tabular}
\end{table}

\section{Prompt and Schema Templates}
\label{app:prompts}

Each evaluation setting specifies a task instruction, input fields, and an ordered set of candidate labels with descriptions. Single-text tasks use a \texttt{text} field, whereas NLI uses separate \texttt{premise} and \texttt{hypothesis} fields. Gold labels and example identifiers are excluded from model inputs. Figure~\ref{fig:prompt-schema-example} provides a schematic illustration using sentiment classification. The task specification is adapted to each model's inference interface. JEV, LAYA, and Quyet use typed-choice questions with \texttt{instructions} and \texttt{criteria}, while GLiNER uses a classification schema with label descriptions and single-label prediction. We evaluate three label-description configurations: English (EN), Vietnamese (VI), and bilingual English-Vietnamese (EN+VI), with English preceding Vietnamese in the bilingual configuration.

\begin{figure}[t]
\centering
\footnotesize
\setlength{\fboxsep}{6pt}
\fbox{%
\begin{minipage}{\dimexpr\columnwidth-2\fboxsep-2\fboxrule\relax}
\raggedright

\textbf{State}\par
\texttt{text}: \texttt{\{input\}}

\medskip
\textbf{Choice question}\par
\texttt{type}: \texttt{choice}\par
\texttt{instructions}:\par
Classify the overall sentiment expressed in Vietnamese student feedback. Select one label based on the provided definitions.

\medskip
\textbf{Candidate labels}\par
\texttt{negative}: The author expresses dissatisfaction, criticism, disapproval, or a negative evaluation.\par
\smallskip
\texttt{neutral}: The text is factual or descriptive without a clear positive or negative evaluation.\par
\smallskip
\texttt{positive}: The author expresses satisfaction, praise, approval, or a positive evaluation.

\end{minipage}%
}
\caption{Illustrative typed-choice schema for sentiment classification. The instruction is translated into English for readability.}
\label{fig:prompt-schema-example}
\end{figure}

\section{Perturbation Implementation Details}
\label{app:perturbation-config}

\subsection{Perturbation Generation}

Teencode and typographical perturbations are generated using \texttt{nom-vn} v0.2.37, while diacritic removal uses the character mapping provided by \texttt{pyvi} v0.1.1. All inputs are normalized to NFC before transformation. Table~\ref{tab:noise-parameters} summarizes the stochastic generator parameters for teencode and typographical noise. OCR and word-segmentation perturbations are disabled in both conditions. Teencode uses dictionary-based slang substitutions, whereas typographical noise combines Telex-related errors, diacritic modifications, confusion-set substitutions, and character-level operations.

\begin{table}[t]
\centering
\footnotesize
\setlength{\tabcolsep}{3pt}
\renewcommand{\arraystretch}{1.10}
\caption{Stochastic perturbation generator parameters.}
\label{tab:noise-parameters}
\begin{tabular}{@{}lrr@{}}
\toprule
Operation / parameter & Teencode & Typo \\
\midrule
\multicolumn{3}{@{}l}{\textit{Token-level operations}} \\
Slang substitution & 0.80 & -- \\
Telex-related errors & -- & 0.08 \\
Full diacritic removal & -- & 0.05 \\
Partial diacritic removal & -- & 0.03 \\
Confusion-set substitution & -- & 0.02 \\
\midrule
\multicolumn{3}{@{}l}{\textit{Character-level thresholds}} \\
Keyboard substitution & -- & 0.015 \\
Adjacent-character swap & -- & 0.025 \\
Character deletion & -- & 0.035 \\
Character duplication & -- & 0.040 \\
\midrule
\texttt{max\_edit\_ratio} & 0.50 & 0.20 \\
\bottomrule
\end{tabular}
\end{table}

Character-level operations are evaluated sequentially using a shared random draw, so the thresholds in Table~\ref{tab:noise-parameters} do not represent independent mutation probabilities. The \texttt{max\_edit\_ratio} parameter limits the internal operation budget for each unprotected segment, with a minimum budget of one, rather than imposing a strict bound on Levenshtein distance. Both stochastic transformations protect URLs, mentions, hashtags, digit-containing expressions, and emoji or symbol sequences. Typographical noise additionally protects selected Vietnamese negation forms, including ``không'', ``chưa'', ``chẳng'', ``chả'', and ``đừng''. Random generation uses a base seed of 42, with seeds derived from example identifiers, input fields, and segment positions.

\subsection{Perturbation Statistics}

Table~\ref{tab:noise-realization-by-dataset} reports the realized perturbation coverage and character edit burden for each evaluation setting. Coverage denotes the proportion of inputs modified, while edit burden is calculated as the sum of field-level Levenshtein distances divided by the total original input length. Both measures are computed on NFC-normalized strings. For NLI, the edit denominator includes both the premise and hypothesis, including the unchanged premise under typographical perturbations. The statistics cover 13 evaluation settings, comprising 33,350 setting--example pairs. UIT-VSFC is counted separately for topic and sentiment classification.

\begin{table*}[t]
\centering
\footnotesize
\setlength{\tabcolsep}{4pt}
\renewcommand{\arraystretch}{1.08}
\caption{Perturbation coverage and character edit burden by evaluation setting (\%).}
\label{tab:noise-realization-by-dataset}
\begin{tabular}{@{}lr*{6}{r}@{}}
\toprule
\multicolumn{1}{@{}c}{\multirow{2}{*}{\textbf{Dataset}}}
& \multicolumn{1}{c}{\multirow{2}{*}{$\boldsymbol{n}$}}
& \multicolumn{2}{c}{\textbf{Teencode}}
& \multicolumn{2}{c}{\textbf{Typo}}
& \multicolumn{2}{c@{}}{\textbf{No accents}} \\
\cmidrule(lr){3-4}
\cmidrule(lr){5-6}
\cmidrule(l){7-8}
& & \multicolumn{1}{c}{Changed}
& \multicolumn{1}{c}{Edit}
& \multicolumn{1}{c}{Changed}
& \multicolumn{1}{c}{Edit}
& \multicolumn{1}{c}{Changed}
& \multicolumn{1}{c@{}}{Edit} \\
\midrule

\multicolumn{8}{@{}l}{\textit{Topic classification}} \\
NEU-ESC
& 6{,}613 & 68.30 & 3.57 & 90.20 & 7.42 & 99.17 & 20.42 \\
UIT-VSFC
& 3{,}166 & 42.61 & 2.20 & 90.24 & 7.49 & 99.87 & 20.44 \\
\midrule

\multicolumn{8}{@{}l}{\textit{Emotion recognition}} \\
UIT-VSMEC
& 693 & 71.72 & 5.05 & 87.73 & 7.73 & 99.86 & 21.62 \\
\midrule

\multicolumn{8}{@{}l}{\textit{Hate speech detection}} \\
ViHSD
& 6{,}680 & 49.03 & 3.30 & 78.95 & 7.23 & 91.53 & 18.99 \\
UIT-ViCTSD
& 1{,}000 & 73.90 & 2.99 & 95.00 & 7.55 & 99.90 & 21.24 \\
UIT-ViOCD
& 549 & 75.59 & 3.83 & 91.07 & 7.55 & 99.82 & 21.49 \\
\midrule

\multicolumn{8}{@{}l}{\textit{Sentiment analysis}} \\
UIT-VSFC
& 3{,}166 & 43.02 & 2.26 & 89.70 & 7.47 & 99.87 & 20.44 \\
\midrule

\multicolumn{8}{@{}l}{\textit{Intent detection}} \\
PhoATIS
& 893 & 51.96 & 1.66 & 94.96 & 7.35 & 100.00 & 19.70 \\
ViMQ
& 1{,}000 & 56.20 & 2.73 & 95.60 & 7.08 & 100.00 & 20.47 \\
\midrule

\multicolumn{8}{@{}l}{\textit{Natural language inference}} \\
ViNLI
& 2{,}991 & 80.51 & 1.62 & 98.90 & 3.16 & 100.00 & 19.78 \\
ViANLI
& 1{,}000 & 76.40 & 1.55 & 98.40 & 2.74 & 100.00 & 19.69 \\
ViMedNLI
& 1{,}422 & 69.48 & 2.43 & 82.49 & 1.83 & 100.00 & 20.51 \\
vnNLI
& 4{,}177 & 77.47 & 1.11 & 99.64 & 3.32 & 100.00 & 14.31 \\
\bottomrule
\end{tabular}
\end{table*}

\begin{table*}[t]
\centering
\footnotesize
\setlength{\tabcolsep}{3.5pt}
\renewcommand{\arraystretch}{1.15}

\caption{Zero-shot macro-$F_1$ on 11 overlapping classification settings. Decision models use English label descriptions, while LLM results are taken from prior zero-shot prompting experiments. Bold and underlined values indicate the highest and second-highest reported scores per setting, respectively. Unreported results are denoted by --.}

\label{tab:decision-llm-comparison}
\resizebox{\textwidth}{!}{%
\begin{tabular}{@{}l*{9}{r}@{}}
\toprule
\multirow{2}{*}{\textbf{Dataset}}
& \multicolumn{5}{c}{\textbf{Large Language Models}}
& \multicolumn{4}{c@{}}{\textbf{Decision Models}} \\
\cmidrule(lr){2-6}
\cmidrule(l){7-10}
& \multicolumn{1}{c}{\shortstack{Gemma-3}}
& \multicolumn{1}{c}{Qwen3}
& \multicolumn{1}{c}{\shortstack{Llama-3.3}}
& \multicolumn{1}{c}{GPT-4o}
& \multicolumn{1}{c}{GPT-4.1}
& \multicolumn{1}{c}{LAYA}
& \multicolumn{1}{c}{GLiNER}
& \multicolumn{1}{c}{Quyet}
& \multicolumn{1}{c@{}}{JEV} \\
\midrule

\multicolumn{10}{@{}l}{\textit{Topic classification}} \\
NEU-ESC
& \textbf{37.32} & 32.12 & 36.49 & 33.24 & 34.15
& 14.79 & 27.10 & 23.08 & \underline{37.31} \\
UIT-VSFC
& 73.60 & 70.27 & 73.11 & 68.99 & \textbf{77.07}
& 46.02 & 48.99 & 63.63 & \underline{74.36} \\
\midrule

\multicolumn{10}{@{}l}{\textit{Emotion recognition}} \\
UIT-VSMEC
& 52.10 & 49.98 & 46.39 & \underline{56.51} & \textbf{57.65}
& 31.89 & 5.10 & 51.76 & 49.99 \\
\midrule

\multicolumn{10}{@{}l}{\textit{Hate speech detection}} \\
ViHSD
& 39.06 & \underline{57.56} & 51.85 & -- & --
& 38.35 & 6.25 & 30.59 & \textbf{59.34} \\
UIT-ViCTSD
& 67.30 & 66.63 & \textbf{70.72} & -- & --
& 58.08 & 50.90 & \underline{69.89} & 69.32 \\
UIT-ViOCD
& \textbf{95.07} & 92.16 & \underline{93.61} & -- & --
& 39.00 & 58.41 & 86.98 & 92.85 \\
\midrule

\multicolumn{10}{@{}l}{\textit{Sentiment analysis}} \\
UIT-VSFC
& 71.09 & 65.62 & 70.66 & 71.37 & \underline{73.14}
& 45.67 & 39.49 & 64.46 & \textbf{73.44} \\
\midrule

\multicolumn{10}{@{}l}{\textit{Intent detection}} \\
ViMQ
& 78.71 & \underline{83.60} & \textbf{84.82} & 67.78 & 69.09
& 37.96 & 38.77 & 37.15 & 80.20 \\
\midrule

\multicolumn{10}{@{}l}{\textit{Natural language inference}} \\
ViNLI
& 77.31 & 76.13 & 75.40 & \textbf{83.97} & \underline{83.89}
& 46.66 & 23.63 & 48.48 & 81.00 \\
ViMedNLI
& 76.72 & 74.24 & 76.17 & \textbf{80.57} & \underline{78.64}
& 57.28 & 23.99 & 59.03 & 73.40 \\
vnNLI
& 77.25 & 84.17 & 74.07 & \underline{86.09} & \textbf{87.38}
& 71.79 & 32.19 & 75.24 & 85.90 \\
\bottomrule
\end{tabular}%
}
\end{table*}

\end{document}